\documentclass[11pt,a4paper]{article}
\usepackage{authblk}
\usepackage[hyperref]{ranlp2021}
\usepackage{times}
\usepackage{latexsym}

\usepackage{graphicx}
\usepackage{multirow}
\usepackage{microtype}
\usepackage{booktabs}
\usepackage{enumitem}

\aclfinalcopy 
\def\aclpaperid{198} 

\title{NEREL: A Russian Dataset with Nested Named Entities, \\ Relations and Events}

\author[1]{\bf{Natalia Loukachevitch}} 
\author[2,3]{\bf{Ekaterina Artemova}}
\author[1,4]{\bf{Tatiana Batura}} 
\author[2,5]{\\ \bf{Pavel Braslavski}}
\author[1]{\bf{Ilia Denisov}}
\author[6]{\bf{Vladimir Ivanov}}
\author[9]{\\ \bf{Suresh Manandhar}}
\author[2]{\bf{Alexander Pugachev}}
\author[2,7,8]{\bf{Elena Tutubalina}}

\affil[1]{Lomonosov Moscow State University, Russia}
\affil[2]{HSE University, Russia}
\affil[3]{Huawei  Noah’s Ark lab, Russia}
\affil[4]{Novosibirsk State University, Russia}
\affil[5]{Ural Federal University, Russia}
\affil[6]{Innopolis University, Russia}
\affil[7]{Kazan Federal University, Russia}
\affil[8]{Sber AI, Russia}
\affil[9]{Wiseyak, United States}

\date{}
\begin{document}
\maketitle
\begin{abstract}
In this paper, we present NEREL, a  Russian dataset for named entity recognition and relation extraction. NEREL is significantly larger than existing Russian datasets: to date it contains 56K annotated named entities and 39K annotated  relations. Its important difference from previous datasets is 
annotation of nested named entities, as well as relations within nested entities and at the discourse level. NEREL can facilitate development of novel 
models that can extract relations between nested named entities, as well as relations on both sentence and document levels.   
NEREL also contains the annotation of events involving named entities and their roles in the events. The NEREL collection is available via \url{https://github.com/nerel-ds/NEREL}.
\end{abstract}

\section{Introduction}

Knowledge bases (KBs) encompass a large amount of structured information about real-world entities and their relationships, which is useful in 
many tasks: information retrieval, automatic text summarization, question answering, conversational and recommender systems \cite{k-bert,han-etal-2020-open,huang2020knowledge}. Even the largest knowledge bases are inherently incomplete, but their manual development is time-consuming and expensive.
Automatic population of knowledge bases from large text collections 
is usually broken down into named entity (NE) recognition, relation extraction (RE), and linking entities to a knowledge base. In turn, training and evaluating models addressing these problems require large and high-quality annotated resources. Currently, most of the available resources of this kind are in English. 

\begin{figure*}[h]
    \centering
    \includegraphics[width=0.95\textwidth, height=30mm]{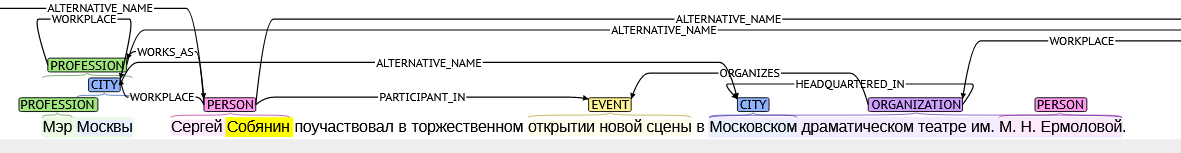}
    \caption{Annotation of the sentence \textit{Moscow Mayor Sergei Sobyanin took part in the grand opening of the new stage of Moscow Ermolova theater} includes nested named entities: \textit{Mayor of Moscow}, \textit{Moscow}, \textit{Mayor}; \textit{Moscow Ermolova Theater}, \textit{Moscow}, \textit{Ermolova}. The intra-entity relations are as follows: \textit{Moscow} is a workplace for \textit{Mayor of Moscow}; \textit{Moscow Ermolova Theater} is headquartered in \textit{Moscow}. \textit{Grand opening of the new stage} is annotated as an event. One can also see \textsc{alternative\_name} relations linking \textit{Moscow} and \textit{Moscow\textsubscript{adj}} within the sentence and \textit{Moscow Mayor},  \textit{Sergei Sobyanin} with other mentions in neighboring sentences.} 
    \label{fig:example}
\end{figure*} 


In this paper, we present NEREL (\textbf{N}amed \textbf{E}ntities and \textbf{REL}ations), a new Russian dataset with annotated named entities and relations.
In developing the annotation schema, we aimed to accommodate recent advances in information extraction methods and datasets.
In particular, 
nested named entities and relations within named entities are annotated in NEREL.
Both of these provide a richer and more complete annotation compared with a flat annotation scheme. 
Current datasets with nested named  entities \cite{ringland2019nne, benikova2014nosta} are not annotated with relations. Therefore, most state-of-the-art relation extraction models~\cite{joshi-etal-2020-spanbert,alt2019improving} do not work with relations between nested and overlapping entities. NEREL aims to address these deficiencies with the addition of nested named entities and relations within nested entities. 

Secondly, NEREL relations are annotated across sentence boundaries at the discourse level allowing for more realistic information extraction experiments.
Figure~\ref{fig:example} illustrates annotation of nested entities, relations between overlapping entities, as well as cross-sentence relations on a sample NEREL sentence.

Finally, NEREL provides annotation for factual events (such as meetings, negotiations, incidents, etc.) involving named entities and their roles in the events.
Future versions of the dataset can easily expand the current inventory of entities and relations. 

NEREL is the largest dataset for Russian annotated with named entities and relations. NEREL features 29 entity and 49 relation types. At the time of writing the dataset contains 56K entities and 39K relations annotated in 900+ Russian Wikinews documents. 
    


In the rest of the paper, we describe the principles behind dataset building process. We also report dataset statistics and provide baseline results for several models. These results indicate that there is a room for improvements. The NEREL collection is freely available.

\section {Related Work}

Table~\ref{tab:datasets} summarizes most important datasets in the context of NEREL development and provides references to their descriptions.



\subsection{Datasets for NER}

\begin{table*}[t]

    \small
    \begin{tabular}{clcccc}
    \toprule \noalign{\smallskip}
    & Dataset  &  Lang & \#NE inst.    & Max & \#Rel inst. \\
    & & & (Types) & Depth & (Types)\\
   \noalign{\smallskip}\midrule\noalign{\smallskip}
    1 & CoNLL03~\cite{conll03} & en & 34.5K (4) & 1 & --\\
    & Ontonotes~\cite{ontonotes} & en & 104K (19) & 1 & --\\
    \noalign{\smallskip}\midrule\noalign{\smallskip}
    2 & ACE2005~\cite{walker2006ace}  & en & 30K (7)    & 6 & 8.3K(6)\\
    & NNE~\cite{ringland2019nne} & en & 279K (114) & 6 & --\\
    & No-Sta-D~\cite{benikova2014nosta}   & de & 41K (12)   & 2 & --\\
    & Digitoday~\cite{ruokolainen2019finnish}  & fi & 19K (6)    & 2 & --\\
    & DAN+~\cite{plank2020dan} & da & 6.4K (4) & 2 & --\\
    \noalign{\smallskip}\midrule\noalign{\smallskip}
    3 & TACRED~\cite{zhang2017tacred} & en & (3) & 1 & 22.8K (42)\\
    & DocRED~\cite{yao2019docred} & en & 132K (6) & 1 & 56K (96)\\
    \noalign{\smallskip}\midrule\noalign{\smallskip}
    4 & Gareev~\cite{Gareev} & ru & 44K (2) & 1 & -- \\
    & Collection3~\cite{mozharova2016two} & ru&26.4K(3)&1&--\\
    & FactRuEval~\cite{FactRuEval2016} & ru & 12K (3) & 2 & 1K (4)\\
    & BSNLP~\cite{piskorski2019second} & ru & 9K (5) & 1 & -- \\
    & RuREBUS~\cite{ivanin2020rurebus} & ru & 121K (5) & 1 & 14.6K (8)\\
    & RURED~\cite{rured} & ru & 22.6K (28) & 1 & 5.3K(34)\\
    \noalign{\smallskip}\midrule\noalign{\smallskip}
    & \textbf{NEREL} (ours)     & ru & 56K (29)   & 6 & 39K (49)\\ \noalign{\smallskip}\bottomrule
    \end{tabular}
   
    \caption{NEREL and its counterparts. Group 1 includes most known datasets with flat entities without relations annotation. Group 2 comprises datasets with nested named entities without or with a small number of relation types. Group 3 includes most known datasets annotated with relations. Group 4 presents Russian datasets for information extraction.}
    \label{tab:datasets}
\end{table*}

Most widely used English datasets for named entity recognition in general domain are CoNLL03 and OntoNotes. CoNLL03 is annotated with four basic NE types -- persons (\textsc{per}), organizations (\textsc{org}), locations (\textsc{loc}), and other named entities (\textsc{misc}), while OntoNotes comprises annotation of 19 NE types, including numeric and temporal ones. Both datasets feature only flat NE annotations.  


There are several datasets with annotated nested named entities, see Table~\ref{tab:datasets}. 
NNE is the largest corpus of this kind, both in terms of entity types and annotated NE mentions.
NNE provides detailed lexical components such as first and last person's names, units (e.g. \emph{tons}), multipliers (e.g. \emph{billion}), etc. 
These result in six levels of nestedness in the dataset. 

The NoSta-D collection of German Wikipedia articles and online newspapers
is annotated with nested named entities of four main classes. Each class can appear in a nominal (proper noun) form, as a part of a token, or as a derivative (adjective) such as ``{\"o}sterreichischen'' (Austrian).  
The Digitoday corpus for Finnish is annotated with six types of named entities (organization, location, person, product, event, and
date). It permits nested entities with the restriction that
an internal entity cannot be of the same class as its top-level entity. For example, \emph{Microsoft Research} is annotated as a flat entity, without additional annotation of the \emph{Microsoft} entity. 
Both NoSta-D and Digitoday datasets allow at most two levels of nesting within entities. 


Amongst NER datasets in Russian,  RURED \cite{rured} provides the largest number of distinct entities with 28 entity types in the RURED dataset of economic news texts.
RURED annotation scheme of named entities mainly follows the OntoNotes guidelines with addition of extra named entities  (currency, group, family,  country, city, etc).  
Currently, FactRuEval \cite{FactRuEval2016} is the only  Russian dataset annotated with nested named entities with at most 2 levels of nesting.  
In FactRuEval, person mentions (PER) can be subdivided into first/last names,  patronymics,  and  nicknames; while organizations and locations -- into their description/type and names (e.g. \emph{[[Microsoft]\textsubscript{NAME} [Corporation]\textsubscript{TYPE}]\textsubscript{ORG}}). 



\subsection{Datasets for Relation Extraction}


One of the largest datasets for relation extraction is the TACRED dataset \cite{zhang2017tacred}.  Relation annotations within TACRED are constructed by querying \textsc{per} and \textsc{org} entities; the returned sentences are annotated by crowd workers (\textsc{gpe} entities are also annotated, the others are treated as values/strings). 
The dataset consists of 106k sentences with entity mention pairs. Each sentence  is  labeled  with  one  of  41  person-  or organization-oriented relation types, or with a \textsc{no\_relation} tag (Table~\ref{tab:datasets} cites the number of ``positive'' cases).
\citet{alt2020tacred} found that more than 50\% of the examples of the TACRED corpus need to be relabeled to improve the performance of baselines models. 
RURED  \cite{rured} is a Russian language dataset that is similar to TACRED. Several relations for events are added such as the date, place, and 
participants of an event. 
The resulting scheme contains 34 relations. The annotation of relations is mainly within sentences. 

RuREBus corpus \cite{ivanin2020rurebus} consists of strategic planning documents issued by the Ministry of Economic Development of the Russian Federation. The data is annotated with eight specialized relations. 

DocRED \cite{yao2019docred} is another dataset that
is annotated with both named entities and relations. The  dataset includes 96 frequent relation types from Wikidata, the relations are annotated at the document level with significant proportion of relations (40.7\%) is across sentence boundaries.   

FactRuEval   \cite{FactRuEval2016} is a Russian language dataset that includes about 1,000 annotated document-level relations of four types (\textsc{ownership,  occupation, meeting}, and \textsc{deal}).

\subsection{Datasets with Annotated Events}

Existing NER datasets usually contain annotations of named events such as hurricanes, battles, wars, or sports events~\cite{ontonotes,ringland2019nne}. For knowledge graph population tasks, it is useful to extract information about significant entity-oriented factual events such as funerals, weddings, or concerts~\cite{rospocher2016building}. However, such an approach significantly complicates the annotation.  
In previous specialized event annotation efforts, an event is defined as an ``explicit occurrence involving participants''
\cite{song2015light,bies2016comparison,mitamura2015event}. Annotators had to tag an event trigger (word or phrase) consisting of the smallest extent of  text expressing  the  occurrence  of  an  event. 
\citet{mitamura2015event}  presented annotation of so-called ``event nuggets'' that can be discontinuous, for example \textit{found guilty} in the sentence \textit{The court \underline{found} him \underline{guilty}}. 
Additionally, events are annotated with special tags 
indicating  whether  or  not  an event occurred. For example, \textsc{actual} tag is used  when  an event actually happened at a particular place and time. 

According to ACE and Light ERE guidelines \cite{linguistic2014deft,walker2006ace}, only events  of particular types are annotated. The  event categories can be: \textsc{Life,  Business, Conflict, Justice}, and others \cite{song2015light,bies2016comparison,mitamura2015event}. 
In ACE and ERE datasets \cite{aguilar2014comparison} annotated events can be provided with arguments, e.g. \textsc{crime\_arg} or \textsc{sentence\_arg} roles for \textsc{justice} events. There are also universal event attributes, e.g. \textsc{place} and \textsc{time}.

TAC-KBP \cite{aguilar2014comparison,mcnamee2010evaluation} and TACRED~\cite{zhang2017tacred} annotations contain event-related slots, e.g \textsc{charges}.
Such relations can be established between entities, even if the corresponding event is not mentioned explicitly. 

\section {Dataset Annotation}
\subsection{Text Selection}

The NEREL corpus consists primarily of Russian Wikinews articles.\footnote{\url{https://ru.wikinews.org/}} 
Wikinews publishes news stories under Creative Commons License (CC BY 2.5) allowing reuse of the published materials. An additional advantage of Wikinews as a document source is that a subset of entities mentioned in the news are linked to corresponding Wikipedia pages making it useful for linking of annotated NEs to Wikidata. 

To select a subset of Wikinews articles for annotation, we first applied NER trained on RURED~\cite{rured} to the whole Wikinews collection. We focused on articles with high density of automatically detected NEs, paying special attention to NEs associated with persons (e.g. \textsc{person, age}). Articles about persons are important for further relation extraction and provide opportunity for  cross-lingual methods using existing datasets~\cite{walker2006ace,zhang2017tacred}. The extracted articles were inspected manually to balance topics and remove inappropriate documents. Finally, 900+ articles were selected for annotation. At the last step of the selection, we retained texts in the size range 1--5 Kb: 
very short texts provide little context for annotation, while long documents are usually non-coherent (e.g. lists of movies or events).

\subsection{Named Entity Annotation}

To define a list of entity types for NEREL, we started with entities in English OntoNotes~\cite{ontonotes} and  RURED~\cite{rured} datasets. Additionally, we considered entity types available in Stanford named entity recognizer \cite{finkel2005incorporating} and TACRED slots such as \textsc{crime} and \textsc{penalty}. \textsc{Award} and \textsc{disease} were added because of their significant frequency in the gathered collection and importance for personal life. 

We followed the following main principles for annotating  named entities:

\begin{itemize}[topsep=0pt,itemsep=-1ex,partopsep=1ex,parsep=1ex]
    \item  
    The entity annotation  schema should be easily amenable for  further entity linking. 
    \item Annotation of internal entities varies depending on the named entity type.  For example, we do not label numbers within numerical entities such as \textsc{date} or \textsc{money}, because such annotations are not essential for relation extraction and entity linking.    
        \item We annotate nested named entities and  named entities consisting of two disjoint spans, but not intersecting named entities. For example, \textit{deputy chairman}  is not annotated within the span \textit{Deputy Chairman of the State Duma Committee} is annotated as \textit{[Deputy [Chairman of the [[State Duma]\textsubscript{ORG} Committee]\textsubscript{ORG}]\\\textsubscript{PROFESSION}]]\textsubscript{PROFESSION}}, because it intersects with the longer named entity.
    \item Adjectives derived from annotated named entities are also annotated with the same tag. Adjectives occur often as internal entities.  Figure~\ref{fig:example} shows an adjective \textit{moskovskii}  (derived from \textit{Moscow}), indicating  the theater's location. 
    \end{itemize}

Currently, there are 29 entity types in NEREL dataset. Further we describe main groups and specific features of entity annotation.

\textbf{Basic entity  types} comprise \textsc{person}, \textsc{organization}, \textsc{location}, \textsc{facility}, \textsc{geopolitical entities}. The latter are subdivided into \textsc{country}, \textsc{state\_or\_province}, \textsc{city}, and \textsc{district}. We also singled out \textsc{family} entity to have possibility to describe relations between families and their members.

\textbf{Temporal and numerical entities} include  \textsc{number}, \textsc{ordinal}, \textsc{date}, \textsc{time}, \textsc{percent}, \textsc{money}, \textsc{age}. and \textsc{Age} entity is usually not annotated separately from \textsc{date} entities \cite{ontonotes,finkel2005incorporating}, but it has their own relations, and therefore it was singled out.

\textsc{Profession} entity denotes  jobs, positions in various organizations, and professional titles. This entity type is significant for extracting relationships of specific persons \cite{zhang2017tacred,rured,FactRuEval2016}. Both capitalized and lowercased \textsc{professions} are annotated in NEREL in contrast to other works \cite{ringland2019nne,ontonotes}. \textsc{Profession} entity is one of the most frequent entities in NEREL. It can have a quite complicated nested structure, in particular,  longest profession spans include  corresponding workplace organization, which allows for a better description of the person's position (Figure~\ref{fig:example}).

\textbf{Physical object} group of entities includes: \textsc{work\_of\_art}, \textsc{product}, and \textsc{award} entities. In contrast to OntoNotes, we introduced a special \textsc{award} entity type because the structure and relations of \textsc{award} entities are quite different from  \textsc{work\_of\_art} entities, and information about awarding is quite frequent in person-oriented texts. 

In flat named entity annotations, different guidelines can be used for \textsc{product} entities annotation. For example, in OntoNotes \cite{ontonotes}, manufacturer and product should be annotated separately as \textsc{org+product}. The same approach is accepted in the Russian Collection3 \cite{mozharova2016two}. In BSNLP-2019 \cite{piskorski2019second} the manufacturer name  should be included into a longer product name.
In the NEREL dataset, the \textsc{product} entity is annotated as a long span, that can include  manufacturer and number subentities.

\paragraph{\textsc{norp} entities} -- nationalities, religious, or political groups -- are usually capitalized in English but lowercase in Russian. In NEREL, \textsc{nationality} entity comprises mainly the following expressions:
(i)~nouns denoting country citizens such as \textit{ukrainec} (\textit{Ukrainian} as a noun); (ii)~adjectives corresponding to nations in contexts different from authority-related, for example \textit{russkii pisatel} (Russian writer). The same adjectives are annotated as \textsc{country} entity in the context of authorities or the origin of organizations. This decision accounts for  most frequent  relations in both contexts. 

\textbf{Legal entities} (\textsc{law}, \textsc{crime}, and \textsc{penalty}) are significant in person-oriented texts for extraction of relations \cite{zhang2017tacred}  in the contemporary news flow, however such entities are usually not annotated in named entity datasets. For example, the TACRED dataset contains  annotations of the \textsc{charge} relation only. What is more, \textsc{Law} and \textsc{crime} entities can be quite long and specific; they are built from names of organizations, persons, countries, etc. \textsc{Penalty} entities often contain period of the penalty or monetary values of fine imposed. 


\begin{figure}[h]
    \centering
    \includegraphics[width=0.50\textwidth]{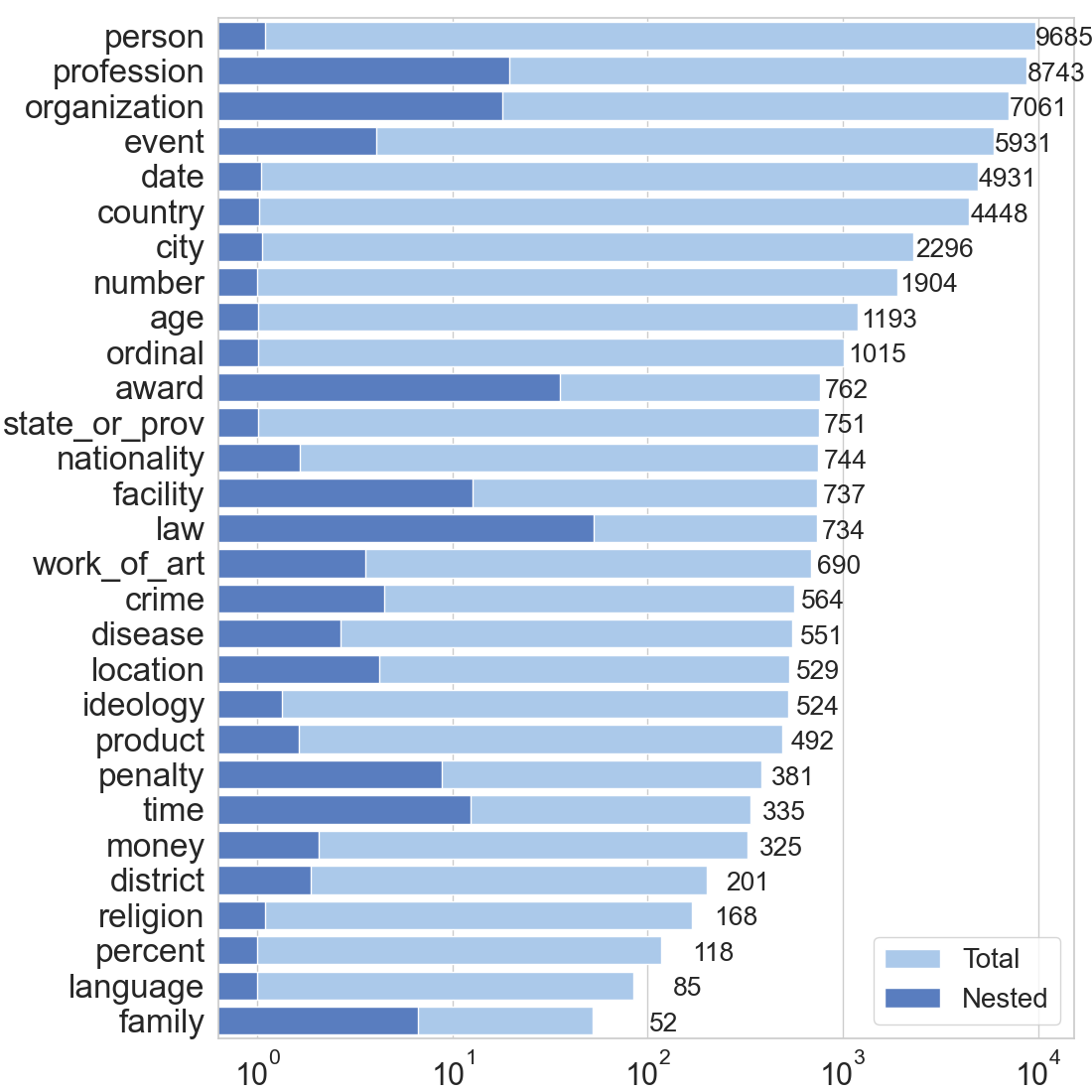}
    \caption{Entity type statistics (log scale). The proportion of nested named entities is shown.}
    \label{fig:NE_stats}
\end{figure}

Entity type frequencies are presented in  Figure~\ref{fig:NE_stats}. As can be seen from the statistics, all but two entity types have at least 100 annotated examples.
Manual annotation of named entities and relations was performed by a single annotator, controlled  by a  moderator. To estimate agreement,
15 documents with about 800 entities were labelled by a moderator (the gold standard) and an annotator. We observed a $F_1$ measure of 92.95 of the annotator's annotation relative to the gold standard, confirming a high level of agreement. Most frequent sources of annotation inconsistencies are as follows: span boundaries of event nuggets, confusing \textsc{facility} and \textsc{organization} entities, confusing  \textsc{event} and \textsc{crime} entities (such as \textit{murders}) or \textsc{event} and \textsc{penalty} entities (such as \textit{arrests}).  \textit{Student} role is often annotated as \textsc{profession} (in spite being a kind of pre-professional title).

NEREL annotations do not contain low-level units as for example the NNE dataset \cite{ringland2019nne} featuring e.g. even  measurement units as separate entities. Annotation of such units is not challenging because it can be performed using  closed word sets. Complex NEREL entities are 
factual \textsc{events} similar to event-nuggets \cite{mitamura2015event} and include \textsc{professions}.
These entity types are extremely useful for further relation extraction. 


\subsection {Events Annotation}
As was mentioned, we annotate both named events (traditionally annotated named sports events, exhibitions, hurricanes, battles, wars, revolutions) \cite{ontonotes,ringland2019nne} and non-named entity-oriented events significant in the news domain. Annotation of non-named events is most similar to event nugget annotation \cite{mitamura2015event}. Event nuggets  can be single words (nouns or verbs) or phrases (noun phrases, verb phrases, or prepositional phrases).  As we annotate entities and relation for knowledge graph population, in the current project mainly factual events, which actually  happened  at  a  particular  place  and time, are labeled. Also we annotate future events with exact dates as if for inclusion in a future schedule.

Main types of annotated factual events are as follows: accidents and deaths (\textit{to crash, to attack}); public actions and ceremonies (\textit{to meet, meeting, summit}); legal actions (\textit{to indict, interrogation, to sentence}); transactions (\textit{to buy, to sell}); appointments and resignations; medical events (\textit{hospitalization, surgical operation}); sports events (\textit{match, final}), etc.


We do not restrict subtypes of entities. We define what we exclude. We exclude from event annotation speech acts and cognitive acts, regular activities,  changes of numerical indicators (for example, prices or import value), victories and defeats.

\subsection {Relation
Annotation}

\begin{table*}[!htp]
    \centering
    \small
    \begin{tabular}{l|l|l|r|r}
    \toprule
Relation & Outer Entity & Inner Entity & Count & \% \\ 
\midrule

\textsc{workplace} & \textsc{profession} & \textsc{organization} & \textsc{1,082} & 19.09 \\ 
\textsc{headquartered\_in} & \textsc{organization} & \textsc{country} & \textsc{846} & 14.93 \\ 
\textsc{workplace} & \textsc{profession} & \textsc{country} & \textsc{669} & 11.81 \\ 
\textsc{headquartered\_in} & \textsc{organization} & \textsc{city} & \textsc{333} & 5.88 \\ 
\textsc{part\_of} & \textsc{organization} & \textsc{organization} & \textsc{281} & 4.96 \\ 
\textsc{headquartered\_in} & \textsc{organization} & \textsc{state\_or\_province} & \textsc{125} & 2.21 \\ 
\textsc{subordinate\_of} & \textsc{profession} & \textsc{profession} & \textsc{116} & 2.05 \\ 
\textsc{workplace} & \textsc{profession} & \textsc{state\_or\_province} & \textsc{116} & 2.05 \\ 
\textsc{part\_of} & \textsc{law} & \textsc{law} & \textsc{111} & 1.96 \\ 
\textsc{ideology\_of} & \textsc{organization} & \textsc{ideology} & \textsc{100} & 1.76 \\ 
\textsc{origins\_from} & \textsc{law} & \textsc{country} & \textsc{100} & 1.76 \\
\midrule


    \end{tabular}
    \caption{The most frequent relation types within nested entities.}
    \label{tab:table_nested}
\end{table*}

Relation types were initially based on TACRED  \cite{zhang2017tacred} and  Russian  RURED \cite{rured} corpora. Further, the list of relations has been corrected and  expanded from the NEREL corpus analysis; corresponding    Wikidata properties were found, when possible.\footnote{Some relations can not have counterparts in Wikidata properties. For example, \textsc{age} and \textsc{age\_died\_at} occur in texts, while Wikidata has only \textit{date of birth (P569)} and  \textit{date of death (P570)} that allow calculating the above mentioned age values.} Names for the relations were selected similar to Wikidata property names. The current set of annotated relation types in the NEREL corpus includes 49 relations.

\begin{figure}[h]
    \centering
    \includegraphics[width=0.50\textwidth]{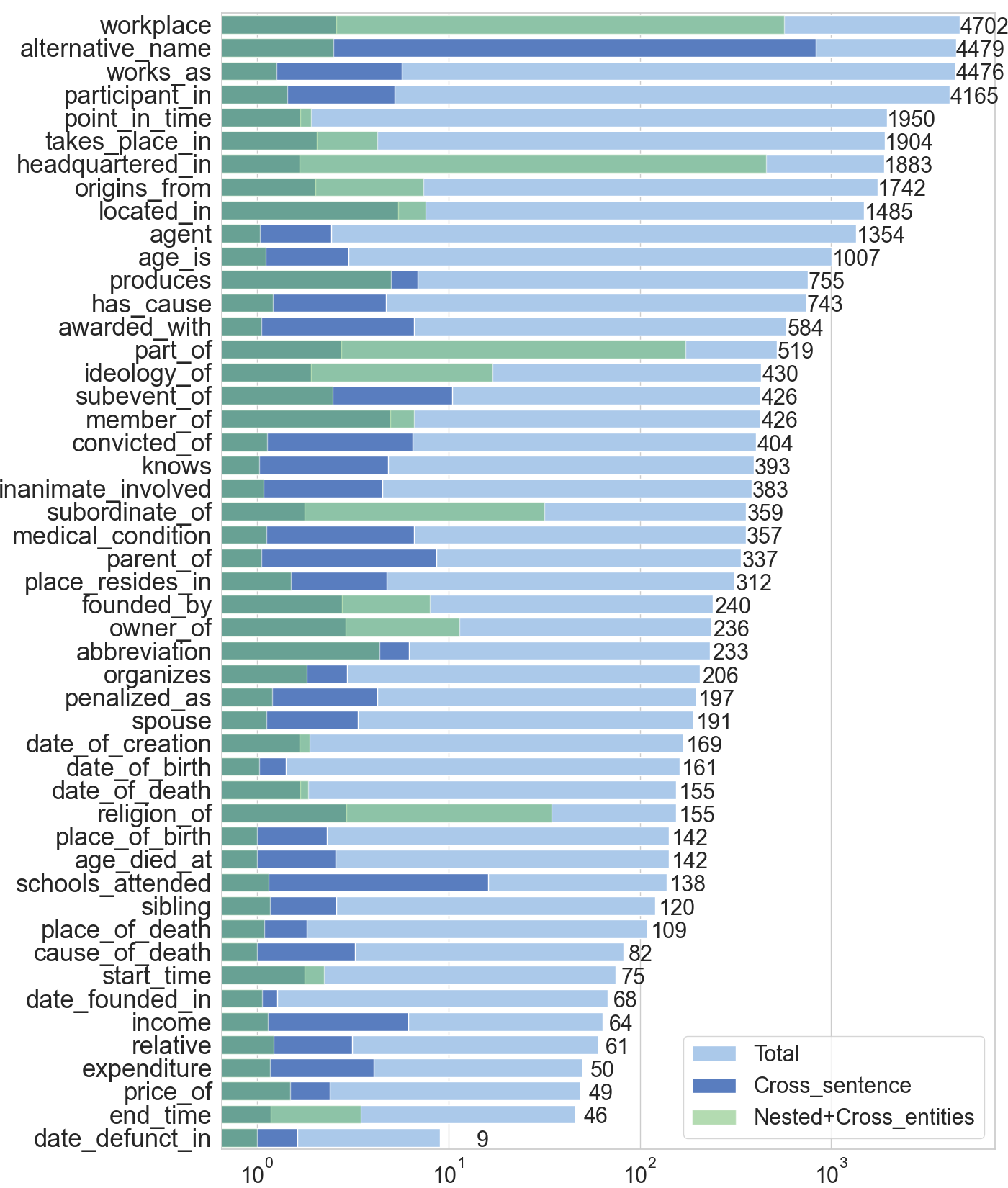}
    \caption{Relation type statistics (log scale). Proportions of cross-sentence relations and  relations involving nestedness of entities are shown.}
    \label{fig:Rel_stats}
\end{figure}

Relations can be subdivided into \textit{person-oriented}, \textit{organization-oriented}, \textit{event-oriented},  and \textit{synonymous} relations (alternative\_name, abbreviation). \textsc{Event}-oriented relations comprise of role relations, temporal relations,  place\_of\_event relation, causal relations, and others.
   Figure~\ref{fig:Rel_stats} shows the distribution of relation frequencies in the NEREL dataset. It can be seen that most relations have at least 50 examples.

All the annotated relations can be subdivided into cross-sentence relations (24\%) and within sentence relations (76\%). Annotated cross-sentence relations make it possible to generate document-level relation extraction, which is important for knowledge graph population from texts.

Among relations within a single sentence, we distinguish three types of relations: 
\begin{itemize}[topsep=0pt,itemsep=-1ex,partopsep=1ex,parsep=1ex]
\item traditional relations between  entities, which are located separately, as \textit{Mayor of Moscow} and \textit{Sergei Sobyanin} in Figure 1 -- further, external relations (52.38\%);  
\item nested relations, i.e. relations within the longest span of a single nested entity, as \textit{Moscow} within \textit{Mayor of Moscow} in Figure 1 (14.75\%);
\item relations crossing entity boundaries, i.e. relations  between an external named entity and an internal entity within a nested named entity (8.87\%) -- further cross-entity relations.

\end{itemize}

The cross-entity relations  can be illustrated as follows: in the sentence \textit{Barack Obama is a member of the Democratic party}, external entity \textit{Barack Obama} has the \textsc{ideology\_of} relation to entity \textit{Democratic}, which is an internal \textsc{ideology} entity in the longer entity \textit{Democratic party}. 

Table~\ref{tab:table_nested} presents the most frequent types of nested named entities connected with nested relations. It can be seen that most nested named entities are professions and professional titles, as well as organization names.



The principles of establishing relations in the NEREL dataset are as follows:
\begin{itemize}[topsep=0pt,itemsep=-1ex,partopsep=1ex,parsep=1ex]
    \item if an entity contains an internal entity of the same type (e.g. \textit{President of Russia~-- President}), all the relations are established with the longer entity. The internal entity helps entity linking if the longer entity is not present in a knowledge base;
    \item all variants of entity names in a single sentence or neighbour sentences are connected with \textsc{alternative\_name} or \textsc{abbreviation} relations, other relations are linked to the closest entity mentions among entities' variants;
    \item cross-sentence relations in neighbouring sentences are annotated with the same detail as in a single sentence; 
    \item relations connecting entities from sentences that are farther than two sentences from each other, 
    should be annotated at least once.
\end{itemize}


\section {Experiments}

We exploit multiple deep learning models, which deliver state-of-the-art results for the English data for two tasks, available at NEREL: (i) nested named entity recognition (NER), (ii) relation extraction. To this end, we subdivided NEREL into train, dev, and test sets --  746/94/93 documents, respectively. 

\subsection{Nested NER} 
We adopted two publicly available models:  Biaffine~\cite{yu2020named} and Pyramid \cite{jue2020pyramid} models with default parameters. Additionally we explored a recently established trend to apply Machine Reading Comprehension (MRC) to nested NER~\cite{li2020unified}.  
The MRC model treats the NER task as extracting answer spans to specialised questions. 
In our case, the questions are dictionary definitions of the words, corresponding to entity types carefully selected from multiple dictionaries. Word representations used with all models are fastText (fT) embeddings
\cite{mikolov2018advances} and pre-trained RuBERT-cased \cite{kuratov2019adaptation}. The latter is utilized in the MRC approach, too. 

Table \ref{tab:Extraction} presents the results of nested NER on the NEREL dataset. The results show, that (i) contextualized BERT-based models outperform models based on static word representations; (ii) the Biaffine model is superior to the Pyramid model; (iii) the results of MRC approach surpass nested NER models' results, most likely, due to the effective usage of additional external information.  However, as the MRC approach treats a single sentence at a time and is than resource-greedy,  the second best solution is still worth consideration. 

\subsection{Relation Extraction} Recent relation extraction models \cite{joshi-etal-2020-spanbert,alt2019improving,han2019opennre} do not support relations between nested named entities or cross-entity relations. These models are tailored to the common test-beds, such as TACRED and DocRED, which do not possess nested named entities, unlike NEREL. Thus we follow the common relation extraction setup and utilize the models to extract  relations between longest entity spans  (i.e. external relations).  To this we adopted three  publicly available models: SpanBERT \cite{joshi-etal-2020-spanbert}, TRE \cite{alt2019improving},  and OpenNRE ~\cite{han2019opennre} with default parameters. The TRE model  is build upon the GPT model \cite{radford2018improving}. Although initially GPT is trained on English web texts, it still has some limited knowledge of Russian, as Russian tokens are present in its vocabulary. The encoders used with SpanBERT and OpenNRE are multilingual BERT and RuBERT. 

\noindent \textbf{Nested relation extraction.} We designed a new model, IntModel, aimed at extraction of nested relations within the longest named entity span. As such relations are contained inside a single entity,  the whole sentence context can be omitted.  To this end, the IntModel classifier inputs the entity features only.  IntModel consists of  a fully-connected layer with the softmax activation, which inputs fastText embeddings of both entities, trainable embeddings of corresponding entity types, and a binary feature showing whether the two entities are nested.   
 
Table  \ref{tab:RelExtraction} presents the results of relation extraction on the NEREL dataset, grouped with respect to three relation types. The results show that (i) overall, in-sentence relations are much easier to extract than the document-level ones; (ii) the monolingual RuBERT provides better results, when compared to the multilingual version and quasi English GPT; (iii) the OpenNRE model is superior to SpanBERT and copes with all three types of relations, (iv) the simplistic IntModel performs on par with more sophisticated models.

\subsection{Discussion} 
Although the preliminary experiments provide with promising results, there is still some room for improvement. Achieved results are comparable to those, published for English datasets, confirming high quality of the collected dataset. At the same time NEREL annotation schema causes difficulties for the current models: all-together nested named entities, combined with diverse relations, require less straightforward approaches, of which machine reading comprehension is one of the promising directions. Detailed error analysis  will help to reveals models' weaknesses and drawbacks.  

\begin{table}[t]
 \begin{center}
   \small
    \begin{tabular}{lccc}
  
    \toprule\noalign{\smallskip}
    Method  &  P    & R &F1 \\
    \noalign{\smallskip}\midrule\noalign{\smallskip}
    Biaffine,  fT&78.84&71.80&75.13\\
    
    Biaffine, RuBERT &\textbf{81.92}&71.54&76.38\\ 
    
    Pyramid, fT&72.70&63.01&67.51\\
    
    Pyramid, RuBERT &77.73&70.97&74.19\\
    \midrule

    MRC&78.70&\textbf{80.24}&\textbf{79.64}\\
    \noalign{\smallskip}\bottomrule
    \end{tabular}
    \caption{Results of nested NER for NEREL}
    \label{tab:Extraction}       
  \end{center}  
   
\end{table}
 
\begin{table}[t]
  \begin{center}
   \small
    \begin{tabular}{llccc}
  
    \toprule\noalign{\smallskip}
    Rel. & Method  &  P    & R & F1 \\
    \noalign{\smallskip}\midrule\noalign{\smallskip}
    \multicolumn{5}{c}{In-sentence relations}\\
    \noalign{\smallskip}\midrule
    \noalign{\smallskip}
     \parbox[t]{2mm}{\multirow{5}{*}{\rotatebox[origin=c]{90}{External}}}  &OpenNRE, mBERT&81.7&81.6&81.7\\
      &OpenNRE, RuBERT&\textbf{85.3}&\textbf{84.6}&\textbf{84.9}\\
      &SpanBERT, mBERT&76.8&75.4&76.1\\
      &SpanBERT, RuBERT&77.4&78.6&78.0\\
      &TRE&66.4&68.1&67.2\\
    
    \noalign{\smallskip}
    \midrule
    \noalign{\smallskip}
     \parbox[t]{2mm}{\multirow{3}{*}{\rotatebox[origin=c]{90}{Nested}}} &OpenNRE, mBERT&74.3&77.7&76.0\\
      &OpenNRE, RuBERT&\textbf{77.8}&\textbf{79.6} &\textbf{78.7}\\
      &IntModel&76.3&72.4&74.3\\
    \noalign{\smallskip}
    \midrule
    \noalign{\smallskip}
    \multicolumn{5}{c}{Document-level relations}\\
    \noalign{\smallskip}\midrule
    \noalign{\smallskip}
    \parbox[t]{2mm}{\multirow{2}{*}{\rotatebox[origin=c]{90}{Doc}}} &OpenNRE, mBERT& 35.7 & 51.2 & 42.1\\
   &OpenNRE, RuBERT& \textbf{52.1} & \textbf{51.3} & \textbf{51.7}\\
    \noalign{\smallskip}\bottomrule
    \end{tabular}
    \caption{Results of relation extraction for NEREL}
    \label{tab:RelExtraction}       
  \end{center}    
   
\end{table}

   


\section {Conclusion}
 We presented a new Russian dataset NEREL annotated with both nested named entity and relations, which is significantly larger than existing Russian datasets. NEREL dataset has several significant distinctive features, including nested named entities, relations over nested named entities, relations on both sentence and discourse level, and events involving named entities.

  NEREL can facilitate development of novel models that address extraction of relations between nested named entities and cross-sentence relation extraction from short texts. NEREL annotation also  allows  relation extraction experiments on both sentence-level and document-level. Nevertheless, NEREL annotations  utilize conventional entity  and relations types, enabling cross-lingual transfer experiments.
  
  Our experiments with baseline models for extraction of entities and relations show that there is room for improvement in both. In the nearest future we plan to enrich the dataset by linking the annotated named entities to Wikidata items. 
 

\section*{Acknowledgments}

The project is supported by the Russian Science Foundation, grant \# 20-11-20166. The experiments were partially carried out on computational resources of HPC facilities at HSE University. We are grateful to Alexey Yandutov and Igor Rozhkov for providing results of their experiments in named entity recognition and relation extraction.  

\pagebreak

\bibliographystyle{acl_natbib}
\bibliography{anthology,ranlp2021}

\begin{thebibliography}{36}
\expandafter\ifx\csname natexlab\endcsname\relax\def\natexlab#1{#1}\fi

\bibitem[{Aguilar et~al.(2014)Aguilar, Beller, McNamee, Van~Durme, Strassel,
  Song, and Ellis}]{aguilar2014comparison}
Jacqueline Aguilar, Charley Beller, Paul McNamee, Benjamin Van~Durme, Stephanie
  Strassel, Zhiyi Song, and Joe Ellis. 2014.
\newblock A comparison of the events and relations across ace, ere, tac-kbp,
  and framenet annotation standards.
\newblock In \emph{Proceedings of the Second Workshop on EVENTS: Definition,
  Detection, Coreference, and Representation}, pages 45--53.

\bibitem[{Alt et~al.(2020)Alt, Gabryszak, and Hennig}]{alt2020tacred}
Christoph Alt, Aleksandra Gabryszak, and Leonhard Hennig. 2020.
\newblock {TACRED} revisited: A thorough evaluation of the {TACRED} relation
  extraction task.
\newblock In \emph{Proceedings of the 58th Annual Meeting of the Association
  for Computational Linguistics}, pages 1558--1569.

\bibitem[{Alt et~al.(2019)Alt, H{\"u}bner, and Hennig}]{alt2019improving}
Christoph Alt, Marc H{\"u}bner, and Leonhard Hennig. 2019.
\newblock Improving relation extraction by pre-trained language
  representations.
\newblock \emph{arXiv preprint arXiv:1906.03088}.

\bibitem[{Benikova et~al.(2014)Benikova, Biemann, and
  Reznicek}]{benikova2014nosta}
Darina Benikova, Chris Biemann, and Marc Reznicek. 2014.
\newblock {NoSta-D} named entity annotation for {German}: Guidelines and
  dataset.
\newblock In \emph{LREC}, pages 2524--2531.

\bibitem[{Bies et~al.(2016)Bies, Song, Getman, Ellis, Mott, Strassel, Palmer,
  Mitamura, Freedman, Ji et~al.}]{bies2016comparison}
Ann Bies, Zhiyi Song, Jeremy Getman, Joe Ellis, Justin Mott, Stephanie
  Strassel, Martha Palmer, Teruko Mitamura, Marjorie Freedman, Heng Ji, et~al.
  2016.
\newblock A comparison of event representations in deft.
\newblock In \emph{Proceedings of the Fourth Workshop on Events}, pages 27--36.

\bibitem[{Finkel et~al.(2005)Finkel, Grenager, and
  Manning}]{finkel2005incorporating}
Jenny~Rose Finkel, Trond Grenager, and Christopher~D Manning. 2005.
\newblock Incorporating non-local information into information extraction
  systems by gibbs sampling.
\newblock In \emph{Proceedings of the 43rd Annual Meeting of the Association
  for Computational Linguistics (ACL’05)}, pages 363--370.

\bibitem[{Gareev et~al.(2013)Gareev, Tkachenko, Solovyev, Simanovsky, and
  Ivanov}]{Gareev}
Rinat Gareev, Maksim Tkachenko, Valery Solovyev, Andrey Simanovsky, and
  Vladimir Ivanov. 2013.
\newblock Introducing baselines for russian named entity recognition.
\newblock In \emph{International Conference on Intelligent Text Processing and
  Computational Linguistics}, pages 329--342.

\bibitem[{Gordeev et~al.(2020)Gordeev, Davletov, Rey, Akzhigitova, and
  Geymbukh}]{rured}
Denis Gordeev, Adis Davletov, Alexey Rey, Galiya Akzhigitova, and Georgiy
  Geymbukh. 2020.
\newblock Relation extraction dataset for the russian language.
\newblock In \emph{Computational Linguistics and Intellectual Technologies:
  Proceedings of the International Conference ``Dialog'' [Komp’iuternaia
  Lingvistika i Intellektual’nye Tehnologii: Trudy Mezhdunarodnoj
  Konferentsii ``Dialog'']}.

\bibitem[{Han et~al.(2020)Han, Cheng, and Wang}]{han-etal-2020-open}
Jiale Han, Bo~Cheng, and Xu~Wang. 2020.
\newblock \href {https://doi.org/10.18653/v1/2020.findings-emnlp.133} {Open
  domain question answering based on text enhanced knowledge graph with
  hyperedge infusion}.
\newblock In \emph{Findings of the Association for Computational Linguistics:
  EMNLP 2020}, pages 1475--1481, Online. Association for Computational
  Linguistics.

\bibitem[{Han et~al.(2019)Han, Gao, Yao, Ye, Liu, and Sun}]{han2019opennre}
Xu~Han, Tianyu Gao, Yuan Yao, Demin Ye, Zhiyuan Liu, and Maosong Sun. 2019.
\newblock {OpenNRE}: An open and extensible toolkit for neural relation
  extraction.
\newblock \emph{EMNLP-IJCNLP 2019}, page 169.

\bibitem[{Hovy et~al.(2006)Hovy, Marcus, Palmer, Ramshaw, and
  Weischedel}]{ontonotes}
Eduard Hovy, Mitchell Marcus, Martha Palmer, Lance Ramshaw, and Ralph
  Weischedel. 2006.
\newblock Ontonotes: The 90\% solution.
\newblock In \emph{Proceedings of the Human Language Technology Conference of
  the NAACL, Companion Volume: Short Papers}, NAACL-Short ’06, page 57–60,
  USA. Association for Computational Linguistics.

\bibitem[{Huang et~al.(2020)Huang, Wu, and Wang}]{huang2020knowledge}
Luyang Huang, Lingfei Wu, and Lu~Wang. 2020.
\newblock Knowledge graph-augmented abstractive summarization with
  semantic-driven cloze reward.
\newblock \emph{arXiv preprint arXiv:2005.01159}.

\bibitem[{Ivanin et~al.(2020)Ivanin, Artemova, Batura, Ivanov, Sarkisyan,
  Tutubalina, and Smurov}]{ivanin2020rurebus}
Vitaly Ivanin, Ekaterina Artemova, Tatiana Batura, Vladimir Ivanov, Veronika
  Sarkisyan, Elena Tutubalina, and Ivan Smurov. 2020.
\newblock Rurebus-2020 shared task: Russian relation extraction for business.
\newblock In \emph{Computational Linguistics and Intellectual Technologies:
  Proceedings of the International Conference “Dialog” [Komp’iuternaia
  Lingvistika i Intellektual’nye Tehnologii: Trudy Mezhdunarodnoj
  Konferentsii “Dialog”]}, Moscow, Russia.

\bibitem[{Joshi et~al.(2020)Joshi, Chen, Liu, Weld, Zettlemoyer, and
  Levy}]{joshi-etal-2020-spanbert}
Mandar Joshi, Danqi Chen, Yinhan Liu, Daniel~S. Weld, Luke Zettlemoyer, and
  Omer Levy. 2020.
\newblock \href {https://doi.org/10.1162/tacl_a_00300} {{S}pan{BERT}: Improving
  pre-training by representing and predicting spans}.
\newblock \emph{Transactions of the Association for Computational Linguistics},
  8:64--77.

\bibitem[{Jue et~al.(2020)Jue, Shou, Chen, and Chen}]{jue2020pyramid}
Wang Jue, Lidan Shou, Ke~Chen, and Gang Chen. 2020.
\newblock Pyramid: A layered model for nested named entity recognition.
\newblock In \emph{Proceedings of the 58th Annual Meeting of the Association
  for Computational Linguistics}, pages 5918--5928.

\bibitem[{Kuratov and Arkhipov(2019)}]{kuratov2019adaptation}
Yuri Kuratov and Mikhail Arkhipov. 2019.
\newblock Adaptation of deep bidirectional multilingual transformers for
  {Russian} language.
\newblock \emph{arXiv preprint arXiv:1905.07213}.

\bibitem[{Li et~al.(2020)Li, Feng, Meng, Han, Wu, and Li}]{li2020unified}
Xiaoya Li, Jingrong Feng, Yuxian Meng, Qinghong Han, Fei Wu, and Jiwei Li.
  2020.
\newblock A unified {MRC} framework for named entity recognition.
\newblock In \emph{Proceedings of the 58th Annual Meeting of the Association
  for Computational Linguistics}, pages 5849--5859.

\bibitem[{{Linguistic Data Consortium}(2014)}]{linguistic2014deft}
{Linguistic Data Consortium}. 2014.
\newblock {DEFT ERE} annotation guidelines: Entities v1.7.

\bibitem[{Liu et~al.(2020)Liu, Zhou, Zhao, Wang, Ju, Deng, and Wang}]{k-bert}
Weijie Liu, Peng Zhou, Zhe Zhao, Zhiruo Wang, Qi~Ju, Haotang Deng, and Ping
  Wang. 2020.
\newblock {K-BERT:} enabling language representation with knowledge graph.
\newblock In \emph{Proceedings of the AAAI Conference on Artificial
  Intelligence}, pages 2901--2908.

\bibitem[{McNamee et~al.(2010)McNamee, Dang, Simpson, Schone, and
  Strassel}]{mcnamee2010evaluation}
Paul McNamee, Hoa~Trang Dang, Heather Simpson, Patrick Schone, and Stephanie~M
  Strassel. 2010.
\newblock An evaluation of technologies for knowledge base population.
\newblock In \emph{LREC}.

\bibitem[{Mikolov et~al.(2018)Mikolov, Grave, Bojanowski, Puhrsch, and
  Joulin}]{mikolov2018advances}
Tom{\'a}{\v{s}} Mikolov, {\'E}douard Grave, Piotr Bojanowski, Christian
  Puhrsch, and Armand Joulin. 2018.
\newblock Advances in pre-training distributed word representations.
\newblock In \emph{Proceedings of the Eleventh International Conference on
  Language Resources and Evaluation (LREC 2018)}.

\bibitem[{Mitamura et~al.(2015)Mitamura, Yamakawa, Holm, Song, Bies, Kulick,
  and Strassel}]{mitamura2015event}
Teruko Mitamura, Yukari Yamakawa, Susan Holm, Zhiyi Song, Ann Bies, Seth
  Kulick, and Stephanie Strassel. 2015.
\newblock Event nugget annotation: Processes and issues.
\newblock In \emph{Proceedings of the The 3rd Workshop on EVENTS: Definition,
  Detection, Coreference, and Representation}, pages 66--76.

\bibitem[{Mozharova and Loukachevitch(2016)}]{mozharova2016two}
Valerie Mozharova and Natalia Loukachevitch. 2016.
\newblock Two-stage approach in russian named entity recognition.
\newblock In \emph{International FRUCT Conference on Intelligence, Social Media
  and Web (ISMW FRUCT)}, pages 1--6.

\bibitem[{Piskorski et~al.(2019)Piskorski, Laskova, Marci{\'n}czuk, Pivovarova,
  P{\v{r}}ib{\'a}{\v{n}}, Steinberger, and Yangarber}]{piskorski2019second}
Jakub Piskorski, Laska Laskova, Micha{\l} Marci{\'n}czuk, Lidia Pivovarova,
  Pavel P{\v{r}}ib{\'a}{\v{n}}, Josef Steinberger, and Roman Yangarber. 2019.
\newblock The second cross-lingual challenge on recognition, normalization,
  classification, and linking of named entities across slavic languages.
\newblock In \emph{Proceedings of the 7th Workshop on Balto-Slavic Natural
  Language Processing}, pages 63--74.

\bibitem[{Plank et~al.(2020)Plank, Jensen, and van~der Goot}]{plank2020dan}
Barbara Plank, Kristian~N{\o}rgaard Jensen, and Rob van~der Goot. 2020.
\newblock Dan+: Danish nested named entities and lexical normalization.
\newblock In \emph{Proceedings of the 28th International Conference on
  Computational Linguistics}, pages 6649--6662.

\bibitem[{Radford et~al.(2018)Radford, Narasimhan, Salimans, and
  Sutskever}]{radford2018improving}
Alec Radford, Karthik Narasimhan, Tim Salimans, and Ilya Sutskever. 2018.
\newblock Improving language understanding by generative pre-training.

\bibitem[{Ringland et~al.(2019)Ringland, Dai, Hachey, Karimi, Paris, and
  Curran}]{ringland2019nne}
Nicky Ringland, Xiang Dai, Ben Hachey, Sarvnaz Karimi, Cecile Paris, and
  James~R Curran. 2019.
\newblock {NNE}: A dataset for nested named entity recognition in english
  newswire.
\newblock In \emph{Proceedings of the 57th Annual Meeting of the Association
  for Computational Linguistics}, pages 5176--5181.

\bibitem[{Rospocher et~al.(2016)Rospocher, van Erp, Vossen, Fokkens, Aldabe,
  Rigau, Soroa, Ploeger, and Bogaard}]{rospocher2016building}
Marco Rospocher, Marieke van Erp, Piek Vossen, Antske Fokkens, Itziar Aldabe,
  German Rigau, Aitor Soroa, Thomas Ploeger, and Tessel Bogaard. 2016.
\newblock Building event-centric knowledge graphs from news.
\newblock \emph{Journal of Web Semantics}, 37:132--151.

\bibitem[{Ruokolainen et~al.(2019)Ruokolainen, Kauppinen, Silfverberg, and
  Lind{\'e}n}]{ruokolainen2019finnish}
Teemu Ruokolainen, Pekka Kauppinen, Miikka Silfverberg, and Krister Lind{\'e}n.
  2019.
\newblock A finnish news corpus for named entity recognition.
\newblock \emph{Language Resources and Evaluation}, pages 1--26.

\bibitem[{Song et~al.(2015)Song, Bies, Strassel, Riese, Mott, Ellis, Wright,
  Kulick, Ryant, and Ma}]{song2015light}
Zhiyi Song, Ann Bies, Stephanie Strassel, Tom Riese, Justin Mott, Joe Ellis,
  Jonathan Wright, Seth Kulick, Neville Ryant, and Xiaoyi Ma. 2015.
\newblock From light to rich ere: annotation of entities, relations, and
  events.
\newblock In \emph{Proceedings of the the 3rd Workshop on EVENTS: Definition,
  Detection, Coreference, and Representation}, pages 89--98.

\bibitem[{Starostin et~al.(2016)Starostin, Bocharov, Alexeeva, Bodrova,
  Chuchunkov, Dzhumaev, Efimenko, Granovsky, Khoroshevsky, Krylova, Nikolaeva,
  Smurov, and Toldova}]{FactRuEval2016}
Anatoly Starostin, Victor Bocharov, Svetlana Alexeeva, Anastasia Bodrova,
  Alexander Chuchunkov, Stanislav Dzhumaev, Irina Efimenko, Dmitry Granovsky,
  Vladimir Khoroshevsky, Irina Krylova, Marina Nikolaeva, Ivan Smurov, and
  Svetlana Toldova. 2016.
\newblock Factrueval 2016: Evaluation of named entity recognition and fact
  extraction systems for russian.
\newblock In \emph{Computational Linguistics and Intellectual Technologies:
  Proceedings of the International Conference “Dialog” [Komp’iuternaia
  Lingvistika i Intellektual’nye Tehnologii: Trudy Mezhdunarodnoj
  Konferentsii “Dialog”]}, pages 702--720.

\bibitem[{Tjong Kim~Sang and De~Meulder(2003)}]{conll03}
Erik~F. Tjong Kim~Sang and Fien De~Meulder. 2003.
\newblock Introduction to the conll-2003 shared task: Language-independent
  named entity recognition.
\newblock In \emph{Proceedings of the Seventh Conference on Natural Language
  Learning at HLT-NAACL 2003 - Volume 4}, CONLL ’03, page 142–147, USA.
  Association for Computational Linguistics.

\bibitem[{Walker et~al.(2006)Walker, Strassel, Medero, and
  Maeda}]{walker2006ace}
Christopher Walker, Stephanie Strassel, Julie Medero, and Kazuaki Maeda. 2006.
\newblock Ace 2005 multilingual training corpus.
\newblock \emph{Linguistic Data Consortium, Philadelphia}, 57:45.

\bibitem[{Yao et~al.(2019)Yao, Ye, Li, Han, Lin, Liu, Liu, Huang, Zhou, and
  Sun}]{yao2019docred}
Yuan Yao, Deming Ye, Peng Li, Xu~Han, Yankai Lin, Zhenghao Liu, Zhiyuan Liu,
  Lixin Huang, Jie Zhou, and Maosong Sun. 2019.
\newblock Docred: A large-scale document-level relation extraction dataset.
\newblock In \emph{Proceedings of the 57th Annual Meeting of the Association
  for Computational Linguistics}, pages 764--777.

\bibitem[{Yu et~al.(2020)Yu, Bohnet, and Poesio}]{yu2020named}
Juntao Yu, Bernd Bohnet, and Massimo Poesio. 2020.
\newblock Named entity recognition as dependency parsing.
\newblock In \emph{Proceedings of the 58th Annual Meeting of the Association
  for Computational Linguistics}, pages 6470--6476.

\bibitem[{Zhang et~al.(2017)Zhang, Zhong, Chen, Angeli, and
  Manning}]{zhang2017tacred}
Yuhao Zhang, Victor Zhong, Danqi Chen, Gabor Angeli, and Christopher~D.
  Manning. 2017.
\newblock Position-aware attention and supervised data improve slot filling.
\newblock In \emph{Proceedings of the 2017 Conference on Empirical Methods in
  Natural Language Processing (EMNLP 2017)}, pages 35--45.

\end{thebibliography}


\end{document}